# Breaking the Language Barrier: Can Direct Inference Outperform Pre-Translation in Multilingual LLM Applications?

Yotam Intrator[*,1]  Matan Halfon[*,1]  Roman Goldenberg[1]  Reut Tsarfaty[2]
Matan Eyal[2]  Ehud Rivlin[1]  Yossi Matias[2]  Natalia Aizenberg[*,1]
[1]Verily Life Sciences, [2]Google Research
{yotami,matanhalfon,nataliea}@google.com

## Abstract

Large language models hold significant promise in multilingual applications. However, inherent biases stemming from predominantly English-centric pre-training have led to the widespread practice of pre-translation, i.e., translating non-English inputs to English before inference, leading to complexity and information loss. This study re-evaluates the need for pre-translation in the context of PaLM2 models (Anil et al., 2023), which have been established as highly performant in multilingual tasks. We offer a comprehensive investigation across 108 languages and 6 diverse benchmarks, including open-end generative tasks, which were excluded from previous similar studies. Our findings challenge the pre-translation paradigm established in prior research, highlighting the advantages of direct inference in PaLM2. Specifically, PaLM2-L consistently outperforms pre-translation in 94 out of 108 languages. These findings pave the way for more efficient and effective multilingual applications, alleviating the limitations associated with pre-translation and unlocking linguistic authenticity.

## 1 Introduction

Large Language Models (LLMs) have become increasingly powerful, leading to their widespread application across various multilingual tasks. However, inherent biases in pre-training data, often heavily skewed towards English, have limited the performance of LLMs on non-English tasks (Srinivasan et al., 2021; Markl, 2022; Tsarfaty et al., 2020). This limitation fueled the standard practice of *pre-translation*, where inputs are translated into English before LLM inference. While bypassing the bias issue, this approach introduces complexities and risks information loss (Nicholas and Bhatia, 2023), impacting both efficiency and effectiveness.

Previous research emphasized the necessity of pre-translation for optimal LLM performance (Shi et al., 2022). However, recent breakthroughs in LLMs trained on massive multilingual datasets, like PaLM2 (Anil et al., 2023), suggest the possibility of overcoming pre-existing biases and enabling direct inference on non-English inputs. This development raises a crucial question: is pre-translation still universally necessary?

While recent studies have extensively explored the impact of pre-translation, their focus has primarily been on discriminative tasks, assessing language understanding (Bandarkar et al., 2023; Etxaniz et al., 2023). The influence of pre-translation on generative capabilities of LLMs, has been largely unexplored. Furthermore, commonly employed performance metrics, when aggregated across languages can be misleading, as outlier results for individual languages can skew the averages and mask the optimal approach for most languages.

This study conducts a comprehensive investigation into the effectiveness of *direct inference* in the source language compared to pre-translation when utilizing PaLM2 models, addressing the aforementioned limitations within existing research. We analyze 108 languages across six different benchmarks, including open-ended tasks to assess the impact of pre-translation on generative abilities. Our findings challenge the established pre-translation paradigm (Ahuja et al., 2023; Shi et al., 2022), highlighting the advantages of direct inference with PaLM2 models. Specifically, PaLM2-L consistently achieves superior performance with direct inference in 94 out of 108 evaluated languages. By revealing PaLM2's superiority with direct inference and offering robust evaluation tools, we aim to inspire further LLM development that transcends pre-translation, paving the way for seamless multilingual communication.

---

[*]These authors contributed equally to this work

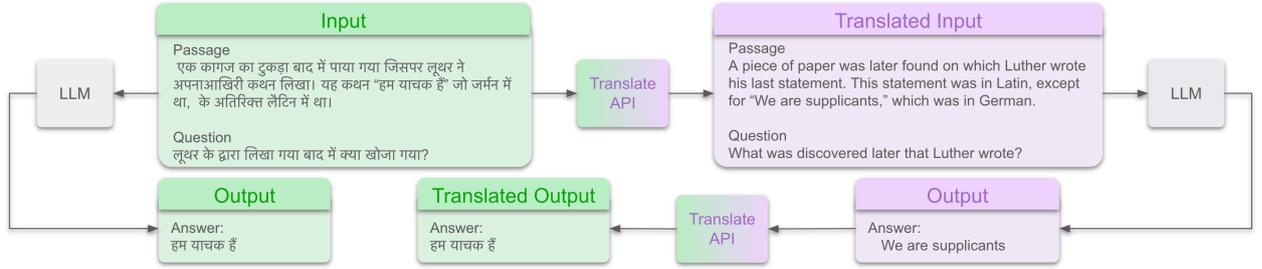

Figure 1: Open-ended question answering piplines for direct (counter clockwise) and pre-translation (clockwise) inference.

## 2 Method

Our primary objective is to determine whether pre-translation remains a necessary practice for optimal LLM performance, specifically focusing on PaLM2, which have been established as highly performant in multilingual tasks (Anil et al., 2023).

In contrast to prior studies, we employ a novel methodology specifically tailored to capture both the nuances of performance across different languages, and asses impact of generative capabilities. To that end, the evaluation incorporates a mixture of closed and open-ended datasets.

While the initial step of pre-translation involves a straightforward translation of the source language to English, evaluating the inference output requires careful consideration. In closed-ended tasks, such as multiple choice question answering (QA) (Bandarkar et al., 2023), the language of the answer is mostly inconsequential. However, open-ended tasks, which assess generation, require the desired answer in the source language, i.e. *evaluation in source language*. This entails an additional post-inference translation step to facilitate evaluation, as illustrated in Figure 1.

Evaluating pre-translation performance for open-ended tasks (text generation), presents unique challenges (Ahuja et al., 2023) stemming from the reliance on lexical evaluation metrics like F1 (Rajpurkar et al., 2016) and ROUGE (Lin, 2004). Addressing these challenges becomes crucial for text summarization and attributive QA datasets utilized in our evaluation.

Attributive QA is a question answering task where the answer relies primarily on information present within the provided context. Here, the ground truth (GT) is a substring of the context, with evaluation metrics measuring the lexical overlap between the predicted answer and the GT. This presents a potential disadvantage for pre-translation, which lacks access to the original context from which the GT was extracted. We address this with a dedicated filtering scheme prescribed in Section 3.

Furthermore, when aggregating and comparing lexical metrics like F1 and ROUGE across languages, inconsistencies arise due to their sensitivity to language morphology. To address this, we propose a complementary *evaluation in English*. We translate both GT and direct inference results to English and calculate lexical metrics in English, regardless of the source language. This way we establish a consistent basis for comparison, eliminating the impact of language-specific variations in the metrics.

Finally, to facilitate a more granular analysis of multiple language performance and address potential bias introduced by a few languages dominating the average, we propose the "Language-Ratio" measure. This measure reports the proportion of languages where one method outperforms the other, offering a more nuanced understanding of relative strengths and weaknesses across languages.

## 3 Experimental Setup

This study utilizes six publicly available multilingual benchmarks for evaluating both discriminative and generative capabilities. For assessing discriminative capabilities, we employ BeleBele (Bandarkar et al., 2023), XCOPA (Ponti et al., 2020), and XStoryCloze (Lin et al., 2021). Generative capabilities are evaluated using XLSum (Hasan et al., 2021), TyDiQA-GP (Clark et al., 2020), and XQuAD (Artetxe et al., 2019). Task types, language counts, and evaluated metrics for each dataset are available in Appendix 7.1.

Two PaLM2 variants are evaluated, PaLM2-S (Bison), and PaLM2-L (Unicorn). Google Trans-

late API[1] is employed for pre-translation. Refer to Appendix 7.2 for detailed prompts used in each dataset.

To achieve a balanced assessment in attributive QA tasks (XQuaD and TyDiQA-GP) we employ filtering. We translate GT from the source language to English and then back to the source language, subsequently discarding any evaluation samples where the translated GT is no longer a substring of the context (details in Appendix 7.5). This approach helps address the inherent bias against pre-translation, discussed in Section 2 by accounting for potential discrepancies in lexical alignment introduced by translation.

## 4 Results and Analysis

### 4.1 Close-ended tasks

Table 1 presents PaLM2's performance on close-ended tasks. We report both average accuracy across languages, in line with prior research, as well as Language-Ratio (i.e. the per language win-rate), providing deeper insights into per-language performance. Notably, both PaLM2-S and PaLM2-L outperform pre-translation approach across all datasets when employing direct inference.

While differences in average accuracy might appear subtle, as exemplified by the 1.1% advantage of PaLM2-S with direct inference on BeleBele, the Language-Ratio highlights a more decisive advantage. In this instance, 74% of languages exhibit superior performance under direct inference. This observation emphasizes the importance of looking beyond average accuracy for a nuanced understanding of model performance across languages.

With this in mind and considering the apparent contradiction with recent findings on the same benchmarks (Ahuja et al., 2023; Bandarkar et al., 2023), we re-examine previously reported results. In (Ahuja et al., 2023) we find that while authors generally concluded that pre-translation yielded superior results across all models, adding Language-Ratio to the results reported on GPT4 reveals a different narrative. Although average accuracy favors pre-translation, in the majority of evaluated languages better performance is achieved with direct inference (see Appendix 7.3 Table 6). While not as conclusive as PaLM2's, this points towards a possible shift in the capabilities of modern language models.

[1]https://cloud.google.com/translate

| Model | Inference | XCOPA | XStoryCloze | BeleBele |
|---|---|---|---|---|
| | | Acc./L% | Acc./L% | Acc./L% |
| PaLM2-S | Pre-translation | 87.3/18.2 | 96.4/30.0 | 76.7/26.0 |
| | Direct | **89.7/81.8** | **96.8/70.0** | **77.8/74.0** |
| PaLM2-L | Pre-translation | 89.6/0.0 | 97.8/0.0 | 84.3/4.8 |
| | Direct | **93.4/90.9** | **99.1/100.0** | **88.4/95.2** |

Table 1: Close-ended tasks pre-translation vs direct inference comparison. Acc. stands for accuracy, L% indicates language-ratio. **Bold** indicates prevailing method for model.

| Model | Inference | XLSum | XQuAD | TyDiQA-GP |
|---|---|---|---|---|
| | | RougeL/L% | F1/L% | F1/L% |
| PaLM2-S | Pre-translation | 23.7/14.6 | 67.2/50.0 | 81.6/12.5 |
| | Direct | **26.8/85.4** | **70.7/50.0** | **83.8/87.5** |
| PaLM2-L | Pre-translation | 25.4/19.5 | 78.7/0.0 | 81.0/25.0 |
| | Direct | **28.0/80.5** | **85.9/100.0** | **83.0/75.0** |

Table 2: Open-ended tasks evaluated in source language. L% indicates language-ratio. **Bold** indicates prevailing method for model.

### 4.2 Open-ended tasks

We evaluated PaLM2 on two generative tasks: question answering (QA) and summarization, employing the two evaluation schemes described in Section 2: (1) *Evaluation in source language* (Table 2) where pre-translated inference is translated to the source language and compared against the original ground truth (GT), and (2) *Evaluation in English* (Table 3) where both pre-translation and direct inference are evaluated against a GT translated to English, requiring an extra step of translation from direct inference to English.

In **Text Summarization**, evaluated through XL-Sum, we find consistently better performance in both models across both evaluation methods, as measured by Rouge-L and Language-Ratio metrics. Furthermore, results from the evaluation in English (Table 3) suggest that if the goal is to summarize content written in the source language into English, translating after inference proves to be a superior to pre-translation.

In **Question Answering**, evaluated via XQuAD and TyDiQA-GP, we analyze two key measures across languages in each benchmark: (1) average F1 score (2) Language-Ratio. While F1 score offers valuable insights, we caution against directly averaging or comparing it across languages due to language-specific morphology influences on the metric. Evaluation in English (Table 3) mitigates this by translating all results to English before calculating F1, leading to more consistent and meaningful comparisons.

| Model | Inference | XLSum | XQuAD | TyDiQA-GP |
|---|---|---|---|---|
| | | RougeL/L% | F1/L% | F1/L% |
| PaLM2-S | Pre-translation | 26.4/26.8 | 70.0/**70.0** | 81.4/12.5 |
| | Direct | **27.8/73.2** | **70.7**/30.0 | **84.7/87.5** |
| PaLM2-L | Pre-translation | 28.3/22.0 | 83.06/10.0 | 81.9/0.0 |
| | Direct | **29.5/78.0** | **86.7/90.0** | **85.2/100.0** |

Table 3: Open-ended evaluated in English. L% indicates language-Ratio. **Bold** indicates prevailing method for model.

We find that PaLM2-L performs consistently better with direct inference compared to pre-translation. PaLM2-S favors direct inference on TyDiQA-GP, however, results on XQuAD are less conclusive. While average F1 scores for direct inference are higher than pre-translation in both evaluation schemes, language-ratio in Evaluation in English suggests an advantage for pre-translation, Table 3). Upon closer inspection of per language F1 in source language (Appendix 7.6 Table 9), we find this discrepancy arises from significant relative improvements in Chinese and Thai with direct inference (30% and 23% respectively), while in most other languages slight performance losses exist.

### 4.3 Language-Focused Analysis

In the analysis above, PaLM2 demonstrated an overall better performance with direct inference, here we wish to inspect in more detail the some of the language specific trends.

PaLM2-L with direct inference consistently outperforms pre-translation in 94 out of 108 evaluated languages (87.04%). However, pre-translation does show consistent superiority in 7 languages: Bambara, Cusco-Collao Quechua, Lingala, Oromo, Punjabi, Tigrinya, and Tsonga (detailed results in Appendix 7.6). Consistency in this context means: consistent results across benchmarks for languages present in multiple datasets. Interestingly, 4 out of the 7 are African languages, with Lingala, the largest, spoken by over 40 million people, suggesting a need for careful examination of African languages when creating multilingual training sets.

The unifying factor across all seven languages however appears to be that they are all low-resource languages (LRL). We categorize LRL as score 2 and below in Joshi et al. (2020b) taxonomy. We conduct an additional analysis focused on LRL (60 languages in total), where direct inference potentially faces greater disadvantage. We inspect lift (Coppock, 2002), i.e, the relative improvement of direct inference over pre-translation, averaged across all benchmarks. Analysis shows that even in LRLs, the majority of languages (over 85%) benefit from direct inference with PaLM2, with lifts exceeding 5% in 63% of languages (see Appendix 7.6 Fig. 4). This reinforces the possibility that the observed performance gaps might indeed have regional origins, highlighting the need for further investigation and potentially tailored approaches for specific language families and regions.

## 5 Conclusions

We performed a comprehensive comparative analysis of direct inference and pre-translation using PaLM2 models on a variety of discriminative and generative tasks across multiple languages. Contrary to prior research, our findings indicate that pre-translation is not universally required and, in fact, direct inference demonstrably improves performance.

In our study we highlight the importance of extending the evaluation scope when assessing the impact of pre-translation. To that end, we propose a set of methods for fair comparison and aggregation of quality metrics across languages. Additionally, we demonstrate the feasibility of evaluation on generative tasks, despite inherent challenges.

Finally, our per-language analysis reveals potential gaps and opportunities for future development in African languages, including those with relatively large speaker populations. These languages appear to be at a disadvantage compared to other low-resource languages, warranting further investigation and targeted efforts to bridge the performance gap.

## 6 Limitations

Our study explores the multilingual landscape, analyzing a diverse range of datasets. However, assessment of open-ended tasks could be improved by using datasets that cover a wider range of languages, similar to the variety in BeleBele, which incorporates over 100 different languages. We identified compelling evidence confirming that direct inference is superior to pre-translation within PaLM2. However, it's important to recognize the potential for similar behavior across a wider spectrum of LLMs.

# References

Kabir Ahuja, Rishav Hada, Millicent Ochieng, Prachi Jain, Harshita Diddee, Samuel Maina, Tanuja Ganu, Sameer Segal, Maxamed Axmed, Kalika Bali, et al. 2023. Mega: Multilingual evaluation of generative ai. *arXiv preprint arXiv:2303.12528*.

Rohan Anil, Andrew M Dai, Orhan Firat, Melvin Johnson, Dmitry Lepikhin, Alexandre Passos, Siamak Shakeri, Emanuel Taropa, Paige Bailey, Zhifeng Chen, et al. 2023. Palm 2 technical report. *arXiv preprint arXiv:2305.10403*.

Mikel Artetxe, Sebastian Ruder, and Dani Yogatama. 2019. On the cross-lingual transferability of monolingual representations. *arXiv preprint arXiv:1910.11856*.

Lucas Bandarkar, Davis Liang, Benjamin Muller, Mikel Artetxe, Satya Narayan Shukla, Donald Husa, Naman Goyal, Abhinandan Krishnan, Luke Zettlemoyer, and Madian Khabsa. 2023. The belebele benchmark: a parallel reading comprehension dataset in 122 language variants. *arXiv preprint arXiv:2308.16884*.

Jonathan H Clark, Eunsol Choi, Michael Collins, Dan Garrette, Tom Kwiatkowski, Vitaly Nikolaev, and Jennimaria Palomaki. 2020. Tydi qa: A benchmark for information-seeking question answering in typologically diverse languages. *Transactions of the Association for Computational Linguistics*, 8:454–470.

David S Coppock. 2002. Why lift?, data modeling and mining. *Information Management Online*, June 21, 2002.

Julen Etxaniz, Gorka Azkune, Aitor Soroa, Oier Lopez de Lacalle, and Mikel Artetxe. 2023. Do multilingual language models think better in english? *arXiv preprint arXiv:2308.01223*.

Tahmid Hasan, Abhik Bhattacharjee, Md Saiful Islam, Kazi Samin, Yuan-Fang Li, Yong-Bin Kang, M Sohel Rahman, and Rifat Shahriyar. 2021. Xl-sum: Large-scale multilingual abstractive summarization for 44 languages. *arXiv preprint arXiv:2106.13822*.

Mandar Joshi, Danqi Chen, Yinhan Liu, Daniel S Weld, Luke Zettlemoyer, and Omer Levy. 2020a. Spanbert: Improving pre-training by representing and predicting spans. *Transactions of the association for computational linguistics*, 8:64–77.

Pratik Joshi, Sebastin Santy, Amar Budhiraja, Kalika Bali, and Monojit Choudhury. 2020b. The state and fate of linguistic diversity and inclusion in the NLP world. In *Proceedings of the 58th Annual Meeting of the Association for Computational Linguistics*, pages 6282–6293, Online. Association for Computational Linguistics.

Chin-Yew Lin. 2004. Rouge: A package for automatic evaluation of summaries. In *Text summarization branches out*, pages 74–81.

Xi Victoria Lin, Todor Mihaylov, Mikel Artetxe, Tianlu Wang, Shuohui Chen, Daniel Simig, Myle Ott, Naman Goyal, Shruti Bhosale, Jingfei Du, et al. 2021. Few-shot learning with multilingual language models. *arXiv preprint arXiv:2112.10668*.

Nina Markl. 2022. Language variation and algorithmic bias: understanding algorithmic bias in british english automatic speech recognition. In *Proceedings of the 2022 ACM Conference on Fairness, Accountability, and Transparency*, pages 521–534.

Gabriel Nicholas and Aliya Bhatia. 2023. Lost in translation: Large language models in non-english content analysis. *arXiv preprint arXiv:2306.07377*.

Edoardo Maria Ponti, Goran Glavaš, Olga Majewska, Qianchu Liu, Ivan Vulić, and Anna Korhonen. 2020. Xcopa: A multilingual dataset for causal commonsense reasoning. *arXiv preprint arXiv:2005.00333*.

Pranav Rajpurkar, Jian Zhang, Konstantin Lopyrev, and Percy Liang. 2016. Squad: 100,000+ questions for machine comprehension of text. *arXiv preprint arXiv:1606.05250*.

Freda Shi, Mirac Suzgun, Markus Freitag, Xuezhi Wang, Suraj Srivats, Soroush Vosoughi, Hyung Won Chung, Yi Tay, Sebastian Ruder, Denny Zhou, Dipanjan Das, and Jason Wei. 2022. Language models are multilingual chain-of-thought reasoners.

Anirudh Srinivasan, Sunayana Sitaram, Tanuja Ganu, Sandipan Dandapat, Kalika Bali, and Monojit Choudhury. 2021. Predicting the performance of multilingual nlp models.

Reut Tsarfaty, Dan Bareket, Stav Klein, and Amit Seker. 2020. From SPMRL to NMRL: What did we learn (and unlearn) in a decade of parsing morphologically-rich languages (MRLs)? In *Proceedings of the 58th Annual Meeting of the Association for Computational Linguistics*, pages 7396–7408, Online. Association for Computational Linguistics.

# 7 Appendix

## 7.1 Datasets

The datasets used in our experiments can be categorized into two types:

- Close-ended task benchmarks (Table 4): These assess discriminative performance. Previous studies have employed these benchmarks for comparing pre-translation and direct inference approaches.

- Open-ended task benchmarks (Table 5): These assess generative abilities. Such benchmarks are commonly used for general multilingual evaluations, but to the best of our knowledge, they have not been previously utilized for assessing the impact of pre-translation.

| Benchmark | Task Type | #Lang | Metrics |
|---|---|---|---|
| BeleBele | Multi-choice QA | 104 | Accuracy/ L% |
| XCOPA | Reasoning | 11 | Accuracy/ L% |
| XStoryCloze | Reasoning | 10 | Accuracy/ L% |

Table 4: Close-ended task datasets used in evaluation and associated metrics. L% stands for Language-Ratio

| Benchmark | Task Type | #Lang | Metrics |
|---|---|---|---|
| XLSum | Summarization | 42 | RougeL/ L% |
| TyDiQA GP | Attributive QA | 8 | F1/ L% |
| XQuAD | Attributive QA | 10 | F1/ L% |

Table 5: Open-ended task datasets used in evaluation and associated metrics. L% stands for Language-Ratio

The total counts of datasets per language as well as associated language Joshi taxonomy (Joshi et al., 2020b) rank are available in Table 13.

In close-ended tasks, the responding model selects the correct answer from a set of predefined options. These tasks typically present a stem, which can be a question, an incomplete statement, or a premise, followed by a list of potential choices. This format restricts the model's response scope, focusing on the retrieval of specific information or confirmation of details rather than encouraging open-ended responses. The following section details the specific types of close-ended tasks and associated benchmark datasets included in this study.

*Commonsense Reasoning* refers to a machine's ability to understand and use everyday knowledge to interpret and respond to natural language. In this study, we employed two commonsense reasoning benchmarks: XCOPA (Ponti et al., 2020) and XStoryCloze (Lin et al., 2021). Both datasets require selecting the correct next sentence, given *premise* among several provided options. In XCOPA, the premise length is not restricted, and two possible answers are provided as next sentence options. XStoryCloze provides a four-sentence premise followed by two possible answers as subsequent sentence options. For XCOPA, Google Translate API is not used as a pre-translated samples are provided for all languages in the dataset.

*Machine Reading Comprehension* assesses a machine's ability to comprehend, interpret, and derive meaning from natural language. We employed BeleBele (Bandarkar et al., 2023), a recently published multilingual aligned dataset, spanning across 122 languages. A task in this dataset contains a question and four candidate answers related to a provided passage. The goal is to select the answer that is most accurate and consistent with the information in the passage. In our evaluation we used the 104 languages supported by Google Translate API.

In open-ended tasks, used to evaluate the models generative abilities, the responding model is given tasks that require it to generate text. Evaluating the correctness of generative tasks mostly relies on lexical overlap, measured by metrics like RougeL or F1, between the generated answer and a GT answer. In our evaluation of open-ended tasks we use two different setups (1) evaluation in source language and (2) evaluation in English, as described in Section 2. Below we detail the types of open-ended tasks included in our study and the associated benchmarks.

*Attributive Question Answering* involves assessing the ability to understand and respond accurately to questions posed in natural language. The most common way of evaluating this automatically is with extractive span tasks, where the respondent is required to form an answer that is a span (i.e. substring) of the provided context. We use XQuAD (Artetxe et al., 2019) and TyDiQA-GP (Clark et al., 2020). These tasks can be satisfied without generation, via span prediction (Joshi et al., 2020a), choosing the beginning and the end of a span within the given context. However, in autoregressive generative models such as LLMs, these tasks serve as proxies for evaluating generation, since the model is prompted to generate the answer token by token. In XQuAD evaluation, we exclude English, and since no development set is provided, we employ zero-shot prompting. In TyDiQA-GP, where a development set is provided, we follow (Anil et al., 2023) and utilize a single-shot. As discussed in Section 3, evaluation of attributive QA requires some level of filtering, for balanced comparison between direct and pre-translation inference. We discussed the details of these in Appendix 7.5.

*Text Summarization*, are tasks aiming to condense a lengthy input text, often a substantial document or article, into a more succinct and focused piece of content that effectively communicates the essential information contained within the original text. We use XLSum (Hasan et al., 2021), an abstractive text summarization benchmark in our evaluation. For calculating Rouge-L on this task, we use an extension discussed in (Anil et al., 2023), that incorporates a SentencePiece tokenizer to handle non-Latin characters. In our construction of prompt, detailed in Section 7.2, we used a single-shot and evaluated it on 41 different languages supported in Google Translate. We followed (Ahuja et al., 2023) and used the first 1,000 samples for each language.

## 7.2 Prompts

### 7.2.1 BeleBele

```
{context}
Question: {question}
Answer A: {possible_answer1}
Answer B: {possible_answer2}
Answer C: {possible_answer3}
Answer D: {possible_answer4}
Correct answer:
```

### 7.2.2 XStoryCloze

```
{sentence1} {sentence2} {sentence3}
{sentence4}
What is a possible continuation for the
story given the following options?
Return either Answer A or Answer B.
Answer A: {possible_answer1}
Answer B: {possible_answer2}
Correct answer:
```

### 7.2.3 XCOPA

```
Premise: {premise}
Question: {question}
Answer A: {possible_answer1}
Answer B: {possible_answer2}
Return either Answer A or Answer B.
Correct answer:
```

### 7.2.4 XQuAD

```
{context}
Question: {question}
The correct answer to the given
question based solely on the
context above is:
```

### 7.2.5 TyDiQA-GP

```
Context: {singleshot_context}
Question: {singleshot_question}
Answer: {singleshot_answer}

Context: {context}
Question: {question}
Answer:
```

### 7.2.6 XLSum

```
Context: {singleshot_context}
Summary: {singleshot_summary}

Context: {context}
Summary:
```

## 7.3 Prior Results GPT4

Analysis of GPT4 performance in previous studies suggests that (1) GPT4 performs better with direct inference on majority of evaluated languages (2) averaging metrics across languages could lead to biased conclusions. In Table 6 we show that despite a higher average accuracy score for the pre-translation, the Language-Ratio metric indicates a preference for direct inference.

| Model | Inference | XCOPA | XStoryCloze |
|---|---|---|---|
| XGLM7.5B | Pre-Translation | **66.3(86.0%)** | **63.6(90.0%)** |
| | Direct | 60.6(14.0%) | 59.9(10.0%) |
| GPT3.5Turbo | Pre-Translation | **81.9**(44.0%) | **93.4**(70.0%) |
| | Direct | 79.1(**56.0%**) | 86.7(20.0%) |
| GPT4-32k | Pre-Translation | **90.6**(22.0%) | **97.0**(30.0%) |
| | Direct | 89.7(**67.0%**) | 96.2(**70.0%**) |

Table 6: The value in parentheses indicates the percentage of the languages the model achieved better performance compared to the other method (pre-translation vs direct). XGLM are taken from (Etxaniz et al., 2023), and GPT from (Ahuja et al., 2023). English was removed from all datasets.

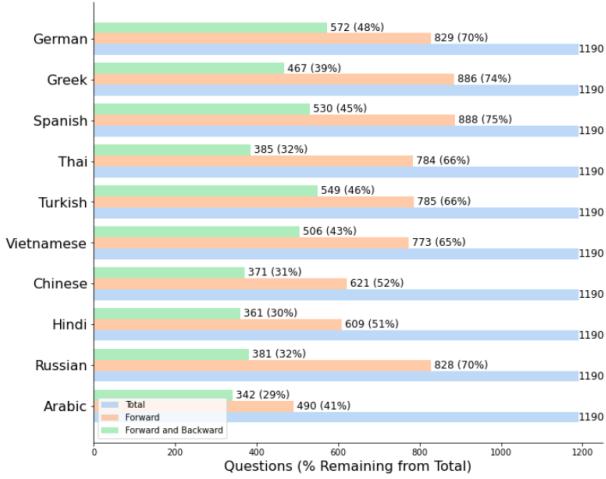

Figure 2: XQuAD filtering.

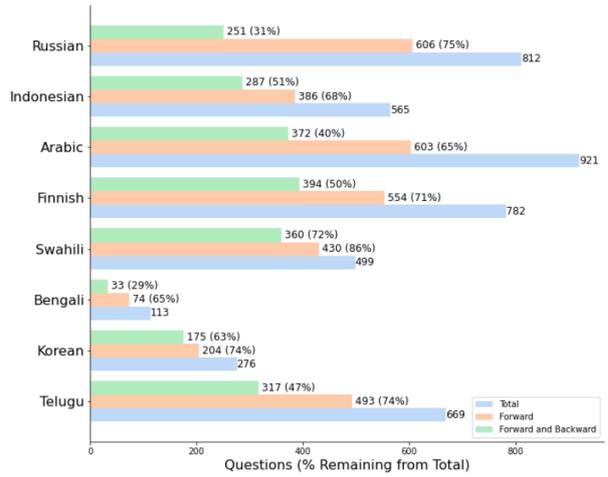

Figure 3: TyDiQA-GP filtering.

## 7.4 Low Resource Languages

We define low-resource languages (LRL) using (Joshi et al., 2020b) taxonomy, specifically considering languages scored at 2 and lower as LRL. Languages which we do not find in the taxonomy at all are also considered low resource. These include languages such as Mesopotamian Arabic, Najdi Arabic, Cusco-Collao Quechua and a few romanized variations. By this categorization 60 out of the 108 languages evaluated in our study are LRL.

To evaluate overall performance of direct inference in comparison to pre-translation in LRL we calculate average of lifts (Coppock, 2002) across benchmarks for each language. We calculate lift as a relative measure of the difference between methods (direct inference relative to pre-translation). This allows us to average findings across different benchmarks, with potentially different quality metrics, for any given language. The lift therefor represents the ratio by which direct inference improves over pre-translation on average across benchmarks for each language.

## 7.5 Filtering

As discussed in Section 3 we employ filtering for a balanced comparison between direct inference and pre-translation in attributive QA.

There are two steps to this process:

- Forward translation filtering - required for both evaluation schemes, i.e. evaluation in source language and evaluation in English.

- Backward translation filtering - only required for evaluation in source language.

Forward translation filtering is when we translate ground truth answer to English, we require it to be a substring of the pre-translated context. This is a definitive pre-requisite for evaluation in English, as otherwise we can not expect the model to be able to predict the correct answer. For evaluation in source language, we also filter by this condition to reduce cases of lexical ambiguity due to synonyms where the pre-translation model may be in a disadvantage. Since prominent concepts tend to have high volume of synonyms across languages, the filtering in English is presumably also helpful for a fair evaluation in source language, though admittedly adding some potential advantage to pre-translation by also removing potentially semantically ambiguous answers to translation.

Backward translation filtering is the follow up step of forward translation filtering, where after translating GT from source language to English, we continue to translate it back to source language and require the back translated GT to be a substring of the original context.

The impact of filtering on the total number of evaluation samples for each language in XQuAD is illustrated in Figure 2. A similar illustration is provided for TyDiQA-GP in Figure 3

## 7.6 Detailed Results

We provide all results across all of the datasets and languages. BeleBele direct vs. pre-translation comparison is in Fig. 5 and Fig. 6 for PaLM2-S and PaLM2-L, respectively. PaLM2-S and PaLM2-L open-ended results are provided in tables 8, 7 and 9.

Figure 4: PaLM2-L average direct inference lift over pre-translate inference on LRL. The majority of languages (over 85%) benefit from direct inference with PaLM2, with lifts exceeding 5% (dashed line) in 63% of languages.

Figure 5: BeleBele PaLM2-L zero shot accuracy comparison.

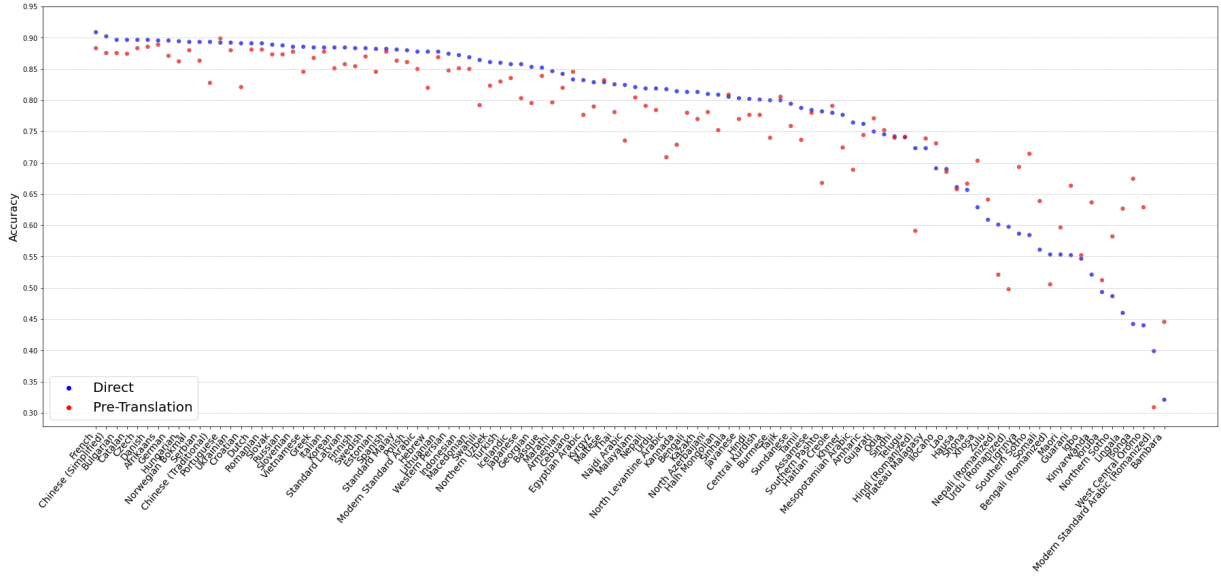

Figure 6: BeleBele PaLM2-S zero shot accuracy comparison.

Table 7: TyDiQA-GP F1 results.

|  | English | | | | Source Language | | | |
|---|---|---|---|---|---|---|---|---|
|  | PaLM2-S | | PaLM2-L | | PaLM2-S | | PaLM2-L | |
| Language | Pre-Translate | Direct | Pre-Translate | Direct | Pre-Translate | Direct | Pre-Translate | Direct |
| Arabic | 78.42 | 80.40 | 79.51 | 80.78 | 79.56 | 79.91 | 79.81 | 79.73 |
| Bengali | 74.12 | 87.40 | 78.27 | 83.98 | 78.42 | 87.95 | 74.38 | 78.55 |
| Finnish | 80.57 | 83.50 | 79.93 | 82.73 | 83.02 | 89.18 | 83.15 | 87.07 |
| Indonesian | 80.59 | 84.96 | 81.98 | 84.20 | 81.70 | 87.29 | 83.84 | 84.96 |
| Korean | 84.42 | 81.75 | 80.19 | 84.2 | 82.95 | 71.05 | 78.17 | 74.19 |
| Russian | 77.35 | 80.77 | 79.39 | 84.14 | 73.77 | 76.35 | 75.60 | 78.15 |
| Swahili | 80.92 | 83.40 | 81.16 | 85.79 | 81.06 | 84.51 | 81.48 | 87.11 |
| Telugu | 94.78 | 95.38 | 94.96 | 95.82 | 92.20 | 94.29 | 91.98 | 94.58 |

Table 8: XLSum RougeL results.

| | English | | | | Source Language | | | |
|---|---|---|---|---|---|---|---|---|
| | PaLM2-S | | PaLM2-L | | PaLM2-S | | PaLM2-L | |
| Language | Pre-Translate | Direct | Pre-Translate | Direct | Pre-Translate | Direct | Pre-Translate | Direct |
| Indonesian | 29.35 | 33.41 | 32.08 | 36.71 | 26.53 | 31.73 | 29.28 | 35.32 |
| Vietnamese | 28.17 | 30.86 | 30.68 | 34.19 | 26.81 | 31.93 | 29.37 | 35.18 |
| Portuguese | 26.47 | 31.22 | 31.16 | 35.21 | 25.40 | 30.89 | 29.80 | 35.13 |
| Japanese | 26.30 | 28.37 | 30.60 | 32.09 | 25.91 | 29.61 | 30.90 | 34.97 |
| Hausa | 29.09 | 29.19 | 30.19 | 35.57 | 27.58 | 28.51 | 28.47 | 34.18 |
| Welsh | 27.16 | 27.46 | 31.99 | 33.25 | 28.30 | 28.24 | 32.91 | 34.08 |
| Swahili | 29.29 | 33.60 | 31.78 | 36.14 | 26.75 | 31.35 | 28.64 | 33.82 |
| Nepali | 26.61 | 31.27 | 30.07 | 33.40 | 25.15 | 31.31 | 27.35 | 33.52 |
| French | 28.90 | 32.34 | 31.74 | 33.76 | 28.23 | 32.20 | 30.66 | 33.49 |
| Persian | 28.06 | 30.00 | 30.90 | 32.06 | 27.00 | 30.53 | 29.27 | 32.57 |
| Chinese | 25.68 | 29.62 | 28.02 | 31.65 | 23.26 | 29.47 | 25.21 | 32.17 |
| Urdu | 28.55 | 31.23 | 29.30 | 32.15 | 27.85 | 31.89 | 28.43 | 32.16 |
| Hindi | 27.84 | 29.78 | 23.42 | 31.64 | 24.88 | 30.15 | 21.82 | 31.85 |
| Sinhala | 29.21 | 29.13 | 29.78 | 34.15 | 25.05 | 26.67 | 25.86 | 31.42 |
| Pashto | 29.22 | 27.19 | 30.63 | 31.09 | 28.93 | 28.42 | 29.90 | 31.04 |
| Arabic | 29.48 | 32.60 | 32.37 | 34.09 | 26.12 | 29.65 | 28.38 | 30.56 |
| Gaelic | 27.43 | 25.88 | 28.12 | 30.69 | 27.02 | 26.46 | 27.87 | 29.73 |
| Turkish | 27.87 | 27.94 | 32.03 | 32.06 | 24.63 | 24.72 | 28.38 | 29.24 |
| Burmese | 25.27 | 28.11 | 25.66 | 26.33 | 23.66 | 31.19 | 24.27 | 29.02 |
| Thai | 22.87 | 24.79 | 24.58 | 27.36 | 18.85 | 24.98 | 21.19 | 28.81 |
| Igbo | 23.74 | 21.77 | 23.94 | 25.28 | 26.62 | 26.69 | 26.87 | 28.40 |
| Russian | 26.92 | 29.96 | 30.95 | 32.36 | 22.04 | 25.76 | 25.49 | 28.30 |
| Ukrainian | 26.88 | 31.11 | 30.22 | 32.86 | 21.79 | 26.07 | 24.83 | 28.27 |
| Somali | 27.68 | 24.00 | 28.50 | 30.73 | 22.94 | 21.40 | 23.47 | 26.46 |
| Amharic | 25.98 | 26.64 | 28.56 | 26.93 | 24.01 | 25.91 | 26.14 | 26.20 |
| Uzbek | 24.66 | 27.44 | 26.21 | 29.26 | 6.96 | 23.75 | 7.19 | 25.82 |
| Chinese (Traditional) | 26.58 | 31.18 | 28.84 | 30.30 | 23.85 | 31.89 | 26.06 | 25.80 |
| Serbian | 24.77 | 25.60 | 30.46 | 29.41 | 21.24 | 21.84 | 25.75 | 25.57 |
| Marathi | 24.30 | 23.96 | 23.85 | 26.61 | 21.13 | 22.75 | 20.60 | 25.41 |
| Gujarati | 23.68 | 24.76 | 25.05 | 27.07 | 21.12 | 23.15 | 22.29 | 25.31 |
| Spanish | 21.01 | 24.18 | 23.85 | 25.94 | 20.10 | 22.87 | 22.58 | 25.10 |
| Punjabi | 26.71 | 22.63 | 27.55 | 23.68 | 27.01 | 23.65 | 27.52 | 24.82 |
| Bengali | 25.32 | 26.48 | 26.57 | 25.80 | 22.46 | 25.14 | 23.43 | 24.78 |
| Tigrinya | 28.87 | 27.79 | 30.63 | 26.10 | 27.40 | 27.70 | 28.12 | 24.58 |
| Yoruba | 24.92 | 23.01 | 24.94 | 28.28 | 23.03 | 23.08 | 23.35 | 24.37 |
| Tamil | 25.88 | 27.21 | 27.48 | 28.20 | 21.00 | 23.72 | 22.39 | 24.21 |
| Oromo | 26.56 | 22.48 | 28.38 | 25.24 | 21.41 | 20.23 | 22.74 | 22.32 |
| Telugu | 23.28 | 24.69 | 20.87 | 23.65 | 19.59 | 22.08 | 18.15 | 20.77 |
| Kyrgyz | 24.16 | 26.57 | 25.10 | 23.14 | 18.66 | 22.25 | 19.51 | 18.00 |
| Azerbaijani | 25.83 | 24.96 | 28.28 | 21.28 | 20.40 | 20.64 | 22.87 | 17.33 |
| Korean | 22.59 | 29.68 | 25.61 | 15.73 | 21.40 | 31.16 | 24.11 | 9.39 |

Table 9: XQuAD F1 results.

| | English | | | | Source Language | | | |
|---|---|---|---|---|---|---|---|---|
| | PaLM2-S | | PaLM2-L | | PaLM2-S | | PaLM2-L | |
| Language | Pre-Translate | Direct | Pre-Translate | Direct | Pre-Translate | Direct | Pre-Translate | Direct |
| Arabic | 72.61 | 66.58 | 83.88 | 87.75 | 72.53 | 63.92 | 82.07 | 88.99 |
| Chinese | 68.71 | 89.6 | 78.96 | 88.44 | 55.92 | 86.79 | 66.63 | 83.97 |
| German | 73.7 | 66.11 | 86.54 | 87.91 | 72.14 | 66.34 | 84.39 | 89 |
| Greek | 71.71 | 65.75 | 85.8 | 87.41 | 71.47 | 67.4 | 83.73 | 88.03 |
| Hindi | 68.22 | 64.66 | 81.98 | 81.8 | 74.87 | 71.14 | 82.76 | 83.98 |
| Russian | 70.88 | 67.55 | 84.13 | 84.6 | 64 | 67.28 | 74.37 | 81.53 |
| Spanish | 70.23 | 68.41 | 85.18 | 86.39 | 67.45 | 68.55 | 81.17 | 85.66 |
| Thai | 66.33 | 81.94 | 79.8 | 89.08 | 60.08 | 79.97 | 73.68 | 88.44 |
| Turkish | 67.43 | 62.7 | 81.7 | 85.77 | 61.2 | 61.15 | 75.34 | 81.49 |
| Vietnamese | 70.66 | 73.41 | 82.6 | 87.47 | 72.21 | 74.89 | 82.89 | 88.1 |

Table 10: XStoryCloze accuracy results.

|  | PaLM2-S | | PaLM2-L | |
| --- | --- | --- | --- | --- |
| Language | Pre-Translation | Direct | Pre-Translation | Direct |
| Basque | 0.960 | 0.952 | 0.980 | 0.997 |
| Russian | 0.981 | 0.971 | 0.990 | 0.995 |
| Chinese | 0.980 | 0.981 | 0.991 | 0.994 |
| Spanish | 0.980 | 0.979 | 0.985 | 0.994 |
| Indonesian | 0.970 | 0.973 | 0.988 | 0.993 |
| Arabic | 0.969 | 0.976 | 0.982 | 0.993 |
| Hindi | 0.965 | 0.975 | 0.978 | 0.991 |
| Burmese | 0.948 | 0.963 | 0.966 | 0.989 |
| Swahili | 0.944 | 0.964 | 0.966 | 0.987 |
| Telugu | 0.938 | 0.950 | 0.972 | 0.979 |

Table 11: XCOPA accuracy results.

|  | PaLM2-S | | PaLM2-L | |
| --- | --- | --- | --- | --- |
| Language | Pre-Translation | Direct | Pre-Translation | Direction |
| Estonian | 0.920 | 0.952 | 0.936 | 0.988 |
| Italian | 0.944 | 0.958 | 0.956 | 0.986 |
| Indonesian | 0.924 | 0.946 | 0.958 | 0.974 |
| Turkish | 0.912 | 0.954 | 0.924 | 0.970 |
| Chinese | 0.950 | 0.944 | 0.966 | 0.970 |
| Vietnamese | 0.920 | 0.928 | 0.942 | 0.964 |
| Tamil | 0.892 | 0.918 | 0.914 | 0.956 |
| Swahili | 0.828 | 0.918 | 0.838 | 0.936 |
| Thai | 0.878 | 0.908 | 0.880 | 0.922 |
| Haitian | 0.848 | 0.844 | 0.848 | 0.920 |
| Cusco-Collao Quechua | 0.596 | 0.604 | 0.694 | 0.694 |

Table 12 – BeleBele accuracy results.

| Language | PaLM2-S | | PaLM2-L | |
|---|---|---|---|---|
| | Pre-Translation | Direct | Pre-Translation | Direct |
| Slovenian | 0.878 | 0.886 | 0.921 | 0.952 |
| Afrikaans | 0.889 | 0.896 | 0.934 | 0.951 |
| Serbian | 0.863 | 0.893 | 0.914 | 0.950 |
| Dutch | 0.881 | 0.891 | 0.923 | 0.949 |
| Standard Latvian | 0.858 | 0.884 | 0.917 | 0.948 |
| Portuguese | 0.899 | 0.892 | 0.947 | 0.948 |
| German | 0.871 | 0.896 | 0.940 | 0.948 |
| French | 0.883 | 0.909 | 0.936 | 0.946 |
| Danish | 0.886 | 0.897 | 0.930 | 0.944 |
| Catalan | 0.874 | 0.897 | 0.921 | 0.944 |
| Vietnamese | 0.846 | 0.886 | 0.909 | 0.943 |
| Lithuanian | 0.869 | 0.878 | 0.909 | 0.942 |
| Romanian | 0.881 | 0.891 | 0.923 | 0.942 |
| Hungarian | 0.862 | 0.894 | 0.922 | 0.942 |
| Ukrainian | 0.880 | 0.892 | 0.911 | 0.942 |
| Chinese (Simplified) | 0.876 | 0.902 | 0.924 | 0.942 |
| Modern Standard Arabic | 0.850 | 0.878 | 0.920 | 0.941 |
| Czech | 0.883 | 0.897 | 0.917 | 0.940 |
| Russian | 0.873 | 0.888 | 0.932 | 0.940 |
| Bulgarian | 0.876 | 0.897 | 0.931 | 0.940 |
| Standard Malay | 0.863 | 0.881 | 0.916 | 0.940 |
| Croatian | 0.821 | 0.891 | 0.917 | 0.939 |
| Greek | 0.868 | 0.884 | 0.918 | 0.938 |
| Norwegian Bokmål | 0.880 | 0.893 | 0.920 | 0.938 |
| Estonian | 0.846 | 0.882 | 0.917 | 0.937 |
| Polish | 0.861 | 0.880 | 0.918 | 0.937 |
| Spanish | 0.878 | 0.882 | 0.927 | 0.937 |
| Slovak | 0.873 | 0.889 | 0.918 | 0.937 |
| Swedish | 0.870 | 0.883 | 0.928 | 0.937 |
| Chinese (Traditional) | 0.828 | 0.893 | 0.907 | 0.937 |
| Italian | 0.878 | 0.884 | 0.917 | 0.936 |
| Swahili | 0.792 | 0.864 | 0.889 | 0.936 |
| Indonesian | 0.851 | 0.872 | 0.919 | 0.934 |
| Finnish | 0.854 | 0.883 | 0.908 | 0.933 |
| Western Persian | 0.848 | 0.874 | 0.892 | 0.931 |
| Georgian | 0.796 | 0.853 | 0.880 | 0.930 |
| Korean | 0.851 | 0.884 | 0.900 | 0.930 |
| Cebuano | 0.846 | 0.833 | 0.902 | 0.929 |
| Northern Uzbek | 0.823 | 0.861 | 0.872 | 0.929 |
| Basque | 0.839 | 0.852 | 0.899 | 0.928 |
| Macedonian | 0.850 | 0.869 | 0.907 | 0.928 |
| Turkish | 0.830 | 0.860 | 0.890 | 0.926 |
| Japanese | 0.803 | 0.858 | 0.899 | 0.926 |
| Maltese | 0.832 | 0.829 | 0.898 | 0.924 |
| Hebrew | 0.820 | 0.878 | 0.882 | 0.921 |
| Kyrgyz | 0.790 | 0.829 | 0.853 | 0.919 |
| Armenian | 0.820 | 0.842 | 0.882 | 0.918 |
| Egyptian Arabic | 0.777 | 0.832 | 0.849 | 0.917 |
| Kannada | 0.729 | 0.814 | 0.883 | 0.914 |
| Urdu | 0.784 | 0.819 | 0.862 | 0.913 |
| Icelandic | 0.836 | 0.858 | 0.882 | 0.913 |
| Central Kurdish | 0.777 | 0.801 | 0.839 | 0.913 |
| Sinhala | 0.809 | 0.806 | 0.879 | 0.912 |
| Tajik | 0.806 | 0.800 | 0.867 | 0.912 |
| Haitian Creole | 0.791 | 0.780 | 0.861 | 0.911 |
| Marathi | 0.797 | 0.847 | 0.862 | 0.908 |
| Odia | 0.752 | 0.746 | 0.841 | 0.907 |
| North Levantine Arabic | 0.709 | 0.818 | 0.806 | 0.907 |
| Malayalam | 0.804 | 0.821 | 0.858 | 0.906 |
| North Azerbaijani | 0.781 | 0.810 | 0.833 | 0.904 |
| Nepali | 0.791 | 0.819 | 0.853 | 0.903 |
| Assamese | 0.780 | 0.784 | 0.847 | 0.902 |
| Gujarati | 0.771 | 0.750 | 0.850 | 0.901 |

Table 12 – BeleBele continued from previous page

| Language | PaLM2-S | | PaLM2-L | |
|---|---|---|---|---|
| | Pre-Translation | Direct | Pre-Translation | Direct |
| Javanese | 0.770 | 0.803 | 0.821 | 0.900 |
| Najdi Arabic | 0.736 | 0.824 | 0.802 | 0.899 |
| Bengali | 0.780 | 0.813 | 0.842 | 0.899 |
| Thai | 0.781 | 0.826 | 0.853 | 0.898 |
| Halh Mongolian | 0.752 | 0.809 | 0.814 | 0.897 |
| Southern Pashto | 0.668 | 0.782 | 0.794 | 0.894 |
| Sindhi | 0.740 | 0.742 | 0.830 | 0.892 |
| Kazakh | 0.770 | 0.813 | 0.829 | 0.891 |
| Amharic | 0.744 | 0.762 | 0.817 | 0.891 |
| Sundanese | 0.759 | 0.794 | 0.839 | 0.889 |
| Hindi | 0.777 | 0.802 | 0.868 | 0.887 |
| Plateau Malagasy | 0.739 | 0.723 | 0.826 | 0.884 |
| Tamil | 0.737 | 0.788 | 0.817 | 0.882 |
| Khmer | 0.724 | 0.777 | 0.819 | 0.881 |
| Burmese | 0.740 | 0.800 | 0.806 | 0.878 |
| Ilocano | 0.731 | 0.691 | 0.861 | 0.869 |
| Hindi (Romanized) | 0.591 | 0.723 | 0.657 | 0.867 |
| Mesopotamian Arabic | 0.689 | 0.764 | 0.738 | 0.860 |
| Telugu | 0.741 | 0.741 | 0.817 | 0.856 |
| Hausa | 0.658 | 0.661 | 0.748 | 0.850 |
| Lao | 0.686 | 0.690 | 0.784 | 0.836 |
| Xhosa | 0.703 | 0.629 | 0.776 | 0.830 |
| Southern Sotho | 0.714 | 0.584 | 0.787 | 0.826 |
| Somali | 0.639 | 0.561 | 0.743 | 0.819 |
| Shona | 0.667 | 0.657 | 0.747 | 0.814 |
| Zulu | 0.641 | 0.609 | 0.721 | 0.813 |
| Urdu (Romanized) | 0.498 | 0.598 | 0.800 | 0.811 |
| Bengali (Romanized) | 0.506 | 0.553 | 0.780 | 0.798 |
| Maori | 0.597 | 0.553 | 0.684 | 0.794 |
| Nepali (Romanized) | 0.521 | 0.601 | 0.782 | 0.791 |
| Tigrinya | 0.693 | 0.587 | 0.800 | 0.780 |
| Guarani | 0.663 | 0.552 | 0.724 | 0.778 |
| Kinyarwanda | 0.637 | 0.521 | 0.701 | 0.760 |
| Northern Sotho | 0.582 | 0.487 | 0.657 | 0.759 |
| Modern Standard Arabic (Romanized) | 0.309 | 0.399 | 0.581 | 0.748 |
| Yoruba | 0.512 | 0.493 | 0.610 | 0.730 |
| Igbo | 0.552 | 0.547 | 0.651 | 0.720 |
| West Central Oromo | 0.629 | 0.440 | 0.721 | 0.657 |
| Tsonga | 0.674 | 0.442 | 0.754 | 0.643 |
| Lingala | 0.627 | 0.460 | 0.708 | 0.624 |
| Bambara | 0.446 | 0.321 | 0.508 | 0.402 |

Table 13 – Language Statistics

| Language | Datasets | Joshi Rank | Language | Datasets | Joshi Rank |
|---|---|---|---|---|---|
| Swahili | 5 | 2 | Southern Sotho | 1 | 1 |
| Russian | 5 | 4 | Slovenian | 1 | 3 |
| Indonesian | 5 | 3 | Slovak | 1 | 3 |
| Chinese | 5 | 5 | Sindhi | 1 | 1 |
| Arabic | 5 | 5 | Shona | 1 | 1 |
| Vietnamese | 4 | 4 | Romanian | 1 | 3 |
| Turkish | 4 | 4 | Punjabi | 1 | 2 |
| Thai | 4 | 3 | Polish | 1 | 4 |
| Telugu | 4 | 1 | Odia | 1 | 1 |
| Spanish | 4 | 5 | Norwegian Bokmål | 1 | 1 |
| Hindi | 4 | 4 | Northern Sotho | 1 | 1 |
| Tamil | 3 | 3 | North Levantine Arabic | 1 | 0 |
| Korean | 3 | 4 | Nepali (Romanized) | 1 | 0 |
| Burmese | 3 | 1 | Najdi Arabic | 1 | 0 |
| Bengali | 3 | 3 | Mongolian | 1 | 1 |
| Yoruba | 2 | 2 | Modern Standard Arabic (Romanized) | 1 | 0 |
| Uzbek | 2 | 3 | Mesopotamian Arabic | 1 | 0 |
| Urdu | 2 | 3 | Maori | 1 | 1 |
| Ukrainian | 2 | 3 | Maltese | 1 | 2 |
| Tigrinya | 2 | 2 | Malayalam | 1 | 1 |
| Somali | 2 | 1 | Malay | 1 | 3 |
| Sinhala | 2 | 0 | Malagasy | 1 | 1 |
| Serbian | 2 | 4 | Macedonian | 1 | 1 |
| Portuguese | 2 | 4 | Lithuanian | 1 | 3 |
| Persian | 2 | 4 | Lingala | 1 | 1 |
| Pashto | 2 | 1 | Latvian | 1 | 3 |
| Oromo | 2 | 1 | Lao | 1 | 2 |
| Nepali | 2 | 1 | Kinyarwanda | 1 | 1 |
| Marathi | 2 | 2 | Khmer | 1 | 1 |
| Kyrgyz | 2 | 1 | Kazakh | 1 | 3 |
| Japanese | 2 | 5 | Kannada | 1 | 1 |
| Italian | 2 | 4 | Javanese | 1 | 1 |
| Igbo | 2 | 1 | Ilocano | 1 | 1 |
| Hausa | 2 | 2 | Icelandic | 1 | 2 |
| Haitian | 2 | 2 | Hungarian | 1 | 4 |
| Gujarati | 2 | 1 | Hindi (Romanized) | 1 | 0 |
| Greek | 2 | 3 | Hebrew | 1 | 3 |
| German | 2 | 5 | Guarani | 1 | 1 |
| French | 2 | 5 | Georgian | 1 | 3 |
| Finnish | 2 | 4 | Gaelic | 1 | 0 |
| Estonian | 2 | 3 | Egyptian Arabic | 1 | 3 |
| Chinese (Traditional) | 2 | 1 | Dutch | 1 | 4 |
| Basque | 2 | 4 | Danish | 1 | 3 |
| Azerbaijani | 2 | 1 | Czech | 1 | 4 |

Table 13 – Language Statistics

| Language | Datasets | Joshi Rank | Language | Datasets | Joshi Rank |
|---|---|---|---|---|---|
| Amharic | 2 | 2 | Cusco-Collao Quechua | 1 | 0 |
| Zulu | 1 | 2 | Croatian | 1 | 4 |
| Xhosa | 1 | 2 | Central Kurdish | 1 | 1 |
| Welsh | 1 | 1 | Cebuano | 1 | 3 |
| Urdu (Romanized) | 1 | 0 | Catalan | 1 | 4 |
| Tsonga | 1 | 1 | Bulgarian | 1 | 3 |
| Tajik | 1 | 1 | Bengali (Romanized) | 1 | 0 |
| Swedish | 1 | 4 | Bambara | 1 | 1 |
| Sundanese | 1 | 1 | Assamese | 1 | 1 |
| Southern Sotho | 1 | 1 | Armenian | 1 | 1 |
| Slovenian | 1 | 3 | Afrikaans | 1 | 3 |